\pdfoutput=1
\documentclass[11pt]{article}

\usepackage[]{EACL2023}

\usepackage{times}
\usepackage{latexsym}

\usepackage[T1]{fontenc}
\usepackage[utf8]{inputenc}

\usepackage{microtype}

\usepackage{babel}
\usepackage{url}
\usepackage{color}
\usepackage{makecell}
\usepackage{bbm}
\usepackage{amsmath}
\usepackage{arydshln}
\usepackage{subcaption}
\usepackage{caption}
\usepackage{multirow}
\usepackage{graphicx}
\usepackage{bm}
\usepackage{amsthm}

\usepackage{tikz}
\newcommand*\circled[1]{\tikz[baseline=(char.base)]{
            \node[shape=circle,draw,inner sep=1pt] (char) {#1};}}

\newcommand\extrafootertext[1]{%
    \bgroup
    \renewcommand\thefootnote{\fnsymbol{footnote}}%
    \renewcommand\thempfootnote{\fnsymbol{mpfootnote}}%
    \footnotetext[0]{#1}%
    \egroup
}
\title{Language-Aware Multilingual Machine Translation \\with Self-Supervised Learning}

\author{Haoran Xu$^{\spadesuit}$, Jean Maillard$^{\heartsuit}$, 
Vedanuj Goswami$^{\heartsuit}$\\ [1em]
$^{\spadesuit}$Johns Hopkins University, $^{\heartsuit}$Meta AI\\[1em]
\texttt{hxu64@jhu.edu}\\
\texttt{\{jeanm,vedanuj\}@meta.com}\\[1em]
}

\begin{document}
\maketitle
\extrafootertext{Work done during an internship at Meta AI Research}
\begin{abstract}

Multilingual machine translation (MMT) benefits from cross-lingual transfer but is a challenging multitask optimization problem. This is partly because there is no clear framework to systematically learn language-specific parameters. Self-supervised learning (SSL) approaches that leverage large quantities of monolingual data (where parallel data is unavailable) have shown promise by improving translation performance as complementary tasks to the MMT task. However, jointly optimizing SSL and MMT tasks is even more challenging. In this work, we first investigate how to utilize \textbf{intra-distillation} to learn more \textit{language-specific} parameters and then show the importance of these language-specific parameters. Next, we propose a novel but simple SSL task, \textbf{concurrent denoising}, that co-trains with the MMT task by concurrently denoising monolingual data on both the encoder and decoder. Finally, we apply \textbf{intra-distillation} to this co-training approach. Combining these two approaches significantly improves MMT performance, outperforming three state-of-the-art SSL methods by a large margin, e.g., 11.3\% and 3.7\% improvement on an 8-language and a 15-language benchmark compared with MASS, respectively\footnote{Code is released at \url{https://github.com/fe1ixxu/CD_ID_MMT}.}.
\end{abstract}

\section{Introduction}
\begin{figure}[ht]
    \centering
    \resizebox{0.9\linewidth}{!}{
    \includegraphics[width=7.5cm]{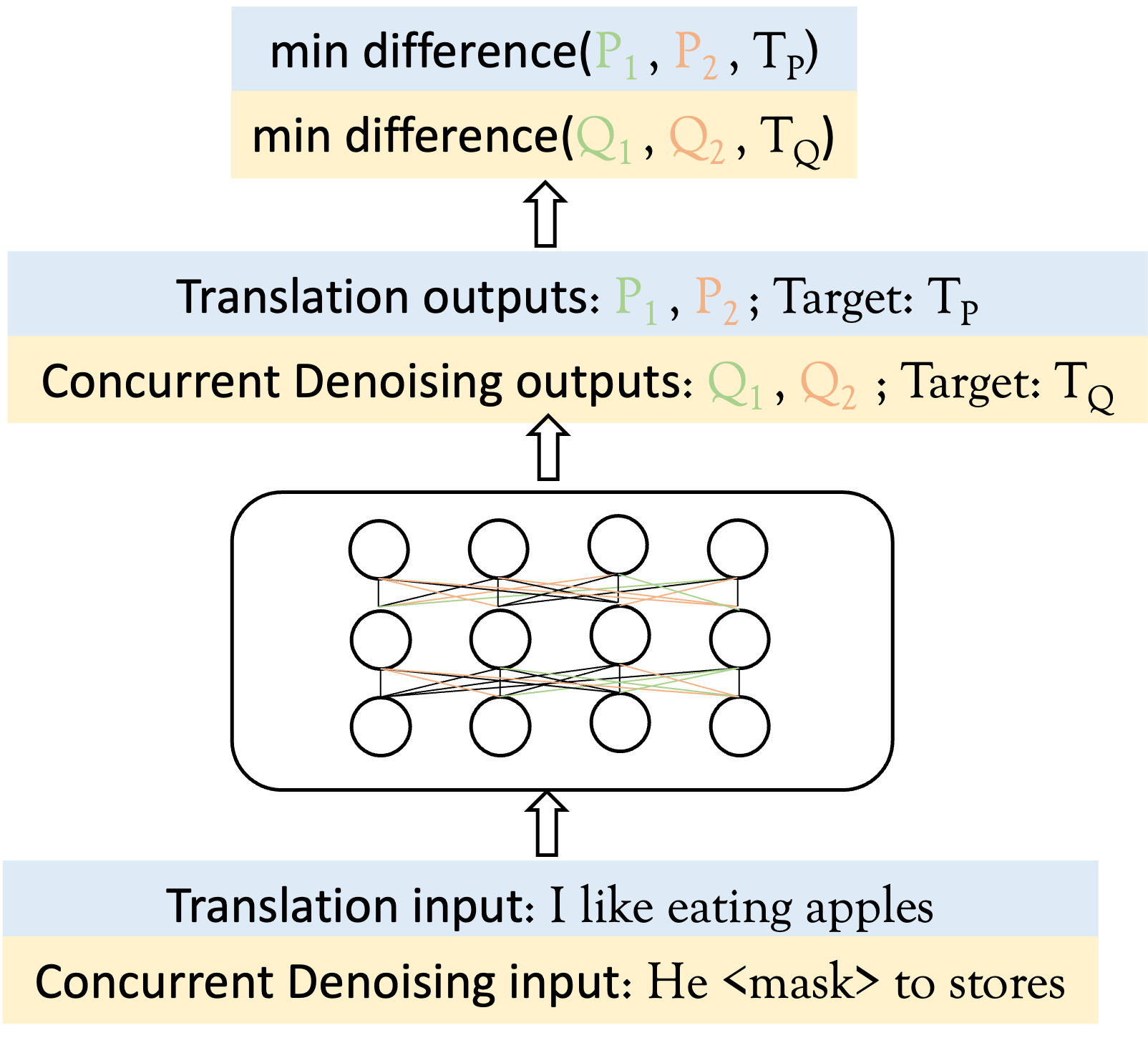}}
    \caption{Concurrent denoising is a complementary task to the MMT task. Both tasks are applied with intra-distillation, where we forward pass model twice for the translation and masked inputs and each time we disable different subsets of parameters (illustrated by different colors). Then, for each task, we not only minimize the difference between the target and two outputs (e.g., minimize difference($P_1$, $T_P$) and difference($P_2$, $T_P$) in the MMT task), we also minimize the difference between two translated outputs as well as two denoised outputs (e.g., minimize difference($P_1$, $P_2$) for MMT).}
    \label{fig:intro}
\end{figure}
Multilingual machine translation (MMT) \citep{aharoni-etal-2019-massively,mmt_challenges} comes with the problem of designing architectures where certain parameters are shared and certain parameters are more language-specific. In order to mitigate negative interference across languages, recent studies have investigated language-specific parameters, including searching for more language-specific parameters \citep{lin-etal-2021-learning}, or adding extra language-specific components to the original model \citep{zhang2021share,nllb}, or even utilizing language-specific pre-trained language models \citep{xu-etal-2021-bert,yarmohammadi-etal-2021-everything}. All these studies indicate the importance of language-specific parameters. \textit{In this work, we first want to encourage parameters to have more language-specific attributes given a fixed model size.}

%Majority of progress in machine translation has been focused on and resulted in improved performance on high-resource languages. 
The difficulty of scaling MMT to low-resource and long-tail languages arises due to the scarcity of abundantly available parallel aligned data. Previous works \citep{nllb,siddhant2022towards,kim2021scalable,wang-etal-2020-multi,mass-mono} try to tackle this by collecting massive amounts of monolingual data and using various types of self-supervised learning (SSL) objectives, such as denoising AutoEncoder (DAE) \citep{mbart} or Masked Sequence to Sequence (MASS) \citep{mass} as auxiliary tasks to co-train with the MMT task, to compensate for the scarcity of parallel data for low-resource languages. \textit{Following this line, we secondly aim to propose a more effective SSL objective.}

With the goal of learning language-aware MMT models and designing more effective SSL methods for MMT, we introduce two approaches. The first approach is \textbf{Intra-Distillation} (ID) \citep{xu2022importance}, which performs a forward pass through the model $K$ times\footnote{We use $K=2$ in this work.}, and in each pass disables a different set of parameters. This enforces consistent contributions between these disabled parameters by minimizing the difference between the $K$ outputs. ID was originally proposed by \citet{xu2022importance} to achieve a balanced parameter contribution in a model. Here, we study the effectiveness of ID in learning language-specific parameters for MMT models. Next, we introduce \textbf{Concurrent Denoising} (CD) which is an auxiliary self-supervised task jointly trained with the MMT task. CD predicts the same masked sentences both on the encoder and decoder side with a shared projection layer to facilitate the consistent understanding between encoder and decoder representations. We show that CD outperforms several state-of-the-art SSL methods for translation. Finally, we apply ID to our co-training scheme to further improve the MMT performance by learning more language-specific parameters. The overall framework is illustrated in Figure \ref{fig:intro} and we summarize our main contributions below.

\begin{itemize}
  \itemsep0em 
    \item We propose a method to quantify the degree of language-specificity of all parameters (Section \ref{sec:preliminary}) and perform a thorough analysis to demonstrate that intra-distillation helps the model learn more language-specific parameters. These parameters contribute more towards a specific language to improve the overall model generalization performance (Section \ref{sec:why_id}).
    \item We propose the \textbf{concurrent denoising} SSL method and demonstrate its improvements over other existing SSL objectives for MMT. Moreover, we introduce a co-training method of MMT and CD with the help of intra-distillation and shows the strong effectiveness of ID in improving MMT+SSL multi-task optimization (Section \ref{sec:method}).
    \item We conduct extensive experiments on a 8-language dataset and a larger 15-language multilingual dataset, and demonstrate that MMT with concurrent denoising and intra-distillation outperforms multiple strong state-of-the-art methods (Section \ref{sec:exp}).

\end{itemize}

\section{Preliminary}
\label{sec:preliminary}
\subsection{Quantify Language-Specific Parameters}
\textbf{Parameter sensitivity} is a measure of the impact on the loss when a specific parameter of a model is zeroed-out. It is widely used in pruning as importance score \citep{ding2019global,molchanov2019importance,lubana2020gradient}. A parameter can express different sensitivities depending on the language of the input data. Those parameters that have high sensitivity to a specific language but low sensitivity to others, are language-specific parameters. We define the $i^\text{th}$ parameter in a model parameterized by $\bm{\Theta}$ as $\theta_i\in\mathbbm{R}$. We further define $\bm{\Theta}_i = [0,\cdots,0,\theta_i,0,\cdots,0]\in\mathbbm{R}^{|\bm{\Theta}|}$ and $\bm{\Theta}_{-i}=[\theta_1,\cdots,\theta_{i-1},0,\theta_{i+1},\cdots,\theta_{|\bm{\Theta}|}]\in\mathbbm{R}^{|\bm{\Theta}|}$. The sensitivity of the $i^\text{th}$ parameter given input batch $b_l$ from language $l$ is formulated as
\begin{equation}
    \mathcal{S}(\theta_i, b_l) = |\mathcal{L}(\bm{\Theta}, b_l) - \mathcal{L}(\bm{\Theta}_{-i}, b_l)|,
    \label{eq:score-define1}
\end{equation}
where $\mathcal{L}(\cdot)$ is the loss function given the input batch and parameters. Then, we use a first-order Taylor decomposition to approximate the sensitivity of any arbitrary parameters. Equation~\ref{eq:score-define1} then becomes
\begin{equation}
    \mathcal{S}(\theta_i, b_l) \approx |\bm{\Theta}_i^T\nabla_{\bm{\Theta}}\mathcal{L}(\bm{\Theta}, b_l)|,
    \label{eq:score-define2}
\end{equation}
where $\nabla_{\bm{\Theta}}\mathcal{L}(\bm{\Theta}, b_l)$ is the gradient of the loss with respect to the model parameters. In our implementation, we randomly pick 500 batches and feed them to the model to retrieve the gradients and compute the average sensitivity. We then have
\begin{equation}
    \mathcal{S}(\theta_i, \mathcal{B}_l) \approx \frac{1}{|\mathcal{B}_l|}\sum_{b_l\in\mathcal{B}_l}|\bm{\Theta}_i^T\nabla_{\bm{\Theta}}\mathcal{L}(\bm{\Theta}, b_l)|,
    \label{eq:score-define3}
\end{equation}
where $\mathcal{B}_l$ is a set containing 500 random $b_l$ batches.

Now, we propose to quantify the degree of language-specificity of $\theta_{i}$ with respect to language $l$ by measuring the relative sensitivity difference between language $l$ and the other languages as
\begin{equation}
    D(\theta_i, l)=\frac{\mathcal{S}(\theta_i, \mathcal{B}_l) - \mathcal{S}(\theta_i, \mathcal{B}_{-l})}{\mathcal{S}(\theta_i, \mathcal{B}_{-l}) + \sigma},
    \label{eq:ls}
\end{equation}
 where $\mathcal{B}_{-l}$ represents the set composed of mixed batches from all training languages except for the language $l$, and $\sigma$ is a very small positive constant\footnote{$\sigma$ is 1e-8 in our implementation.}. The larger $D(\theta_i, l)$ is, the more language-specific $\theta_{i}$ is to language $l$.
 
\subsection{Intra-Distillation}
A model with more balanced parameter sensitivity distribution shows better generalization \citep{sage}. \citet{xu2022importance} propose intra-distillation (ID) as an effective task-agnostic training method, aiming to encourage all parameters to contribute equally, which improves performance when model size is fixed. However we argue that, in the multilingual setting, ID actually helps the model learn more language-specific parameters resulting in improved performance. Given an input batch, ID needs to forward pass the model $K$ times to obtain $K$ outputs and each time a random subset of parameters is zeroed out. The core idea of ID is to minimize the difference of these $K$ outputs to approximate minimizing the contribution gap of the parameters that are zeroed-out, because the $K$ outputs are forced to be the same with different zeroed parameters. Let $\{p_1, \cdots, p_i, \cdots, p_K\}$ denote the $K$ outputs. Note that the outputs are probability distributions in the translation and denoising task. The ID loss is then formulated by the X-divergence \citep{xu2022importance} to minimize the difference of $K$ outputs as
\begin{equation}
\begin{gathered}
    \mathcal{L}_{id} = \frac{1}{K}\sum_{i=1}^K \mathbbm{KL}(p_i\parallel\bar{p}) + \mathbbm{KL}(\bar{p}\parallel p_i) \\
    \text{where }\bar{p} = \frac{1}{K}\sum_{i=1}^Kp_i
    \label{eq:id_loss}
\end{gathered}
\end{equation}

Let the original task loss be $\mathcal{L}_i$ for the $i^\text{th}$ pass. Then, the total loss is a combination of the original task losses and ID loss, given as
\begin{equation}
    \min \frac{1}{K}\sum_{i=1}^K\mathcal{L}_i + \alpha \mathcal{L}_{id}
    \label{eq:id_final_loss}
\end{equation}
where $\alpha$ is a hyper-parameter to control the strength of ID. Similar to \citet{xu2022importance}, we use dropout to simulate \textit{zeroed-out} parameters in all experiments.

Although the explanation for better performance after using ID is that the model parameters become more balanced, it is unclear how parameter contributions to different languages change after applying ID in a multilingual (multitask) setting. For instance, do parameters become more language-agnostic and shareable across all languages, or do they become more language-specific? We investigate this in more details in Section \ref{sec:lg-aware}.

% Interestingly, our findings show that parameters become more language-specific after applying ID, i.e, they tend to contribute more towards a specific language and less to the others (more details in Section \ref{sec:why_id}). This is in contrast to the intuition that more balanced parameter contributions correspond to parameters that are more evenly shared across languages. In fact, learning more language-specific parameters through ID in MMT leads to better performance. This finding aligns with the results of recent studies which investigate language-specific parameters \citep{lin-etal-2021-learning,zhang2021share,nllb}, indicating the importance of language-specific parameters.
%However, the overall parameter contributions are still more balanced as claimed in \citet{xu2022importance}.

\section{Language-Aware MMT Models}
\label{sec:why_id}
In this section, we study how parameters can be prompted to be more language-specific by applying \textbf{intra-distillation}, which improves the model generalization performance. Specifically, certain parameters become more language-specific and tend to contribute more to their specific language and less to others. We demonstrate the importance of language-specific parameters by showing how much they can contribute in pruning experiments. We begin our analysis from a case study on MMT experiments with an 8-language dataset (M8), and then scale up our experiments to 15 languages (M15) with larger data size in Section \ref{sec:exp}. Here, we show results and analysis on  \texttt{xxx}$\rightarrow$\texttt{eng} directions. Similar discussions for \texttt{eng}$\rightarrow$\texttt{xxx} directions are shown in Appendix \ref{app:eng-xx}.

\subsection{Experiments on Intra-Distillation}
\label{sec:why_id_exp}
\paragraph{Dataset and Training}
We train MMT models with and without ID on the M8 dataset\footnote{The languages were selected in order to have a realistic dataset reflecting a specific use case. Multilingual training is crucial for languages that are low-resource, as is the case for many languages of Africa. We chose two different language groupings from the African continent: Benue-Congo languages (Kimbundu, Ganda, Chewa, Swahili, Umbundu, Zulu) and North-Central Atlantic languages (Nigerian Fulfulde, Wolof). While these languages may all belong to the Atlantic-Congo family, this is an extremely large, varied, and under-researched family, with Glottolog recording over 1,400 languoids in it – compare this to under \~590 languoids recorded for the Indo-European family.}. M8 is composed of Nigerian Fulfulde (\texttt{fuv}, 18K parallel sentences), Kimbundu (\texttt{kmb}, 82K), Ganda (\texttt{lug}, 278K), Chewa (\texttt{nya}, 693K), Swahili (\texttt{swh}, 2.1M), Umbundu (\texttt{umb}, 193K), Wolof (\texttt{wol}, 9K) and Zulu (\texttt{zul}, 1.2M). Datasets are extracted from the primary bitext used by the NLLB-200 model \citep{nllb}. For ID, we pass the model twice ($K=2$) considering the computational cost, and set $\alpha$ as 5 suggested by \citet{xu2022importance}. We use \textsc{Flores-200} as our dev and test sets \citep{nllb}. Our model training is based on the Transformer$_\text{big}$ architecture \citep{transformer} with 32K vocabulary jointly trained by SentencePiece \citep{sentencepiece}. We report sacreBLEU scores (\texttt{spm} tokenizer) \citep{sacrebleu}. 

\paragraph{Results}
Following \citet{nllb}, we categorized a language as \textit{low-resource} if there are fewer than 1M parallel sentences, and as \textit{very low-resource} if fewer than 100K (very low-resource is not the subset of low-resource). Otherwise, the language is considered as \textit{high-resource}. We report the average BLEU scores for each of the three categories. In Table \ref{tab:id_vs_regular}, we show that MMT with ID outperforms the regular MMT model by a large margin on all three categorizes by +1.21 BLEU averaged across all languages.

\begin{table}[ht]
\centering
\resizebox{1\linewidth}{!}{
\begin{tabular}{l|cccc}
\hline
Method             & High  & Low   & Very Low & All   \\ \hline
Regular            & 31.70 & 12.57 & 6.92     & 15.94 \\
Intra-Distillation & \textbf{33.30} & \textbf{13.63} & \textbf{8.05}     & \textbf{17.15} \\ \hline
\end{tabular}
}
\caption{M8 results on \texttt{xxx}$\rightarrow$\texttt{eng} comparing regular MMT and MMT with ID. We observe that MMT with ID outperforms regular MMT by a significant margin.}
\label{tab:id_vs_regular}
\end{table}

\subsection{Language-Specific or Language-Agnostic?}
\label{sec:lg-aware}
\begin{figure*}[ht]
     \centering
     \begin{subfigure}[b]{0.7\textwidth}
         \centering
         \resizebox{1\linewidth}{!}{
         \includegraphics[width=\textwidth]{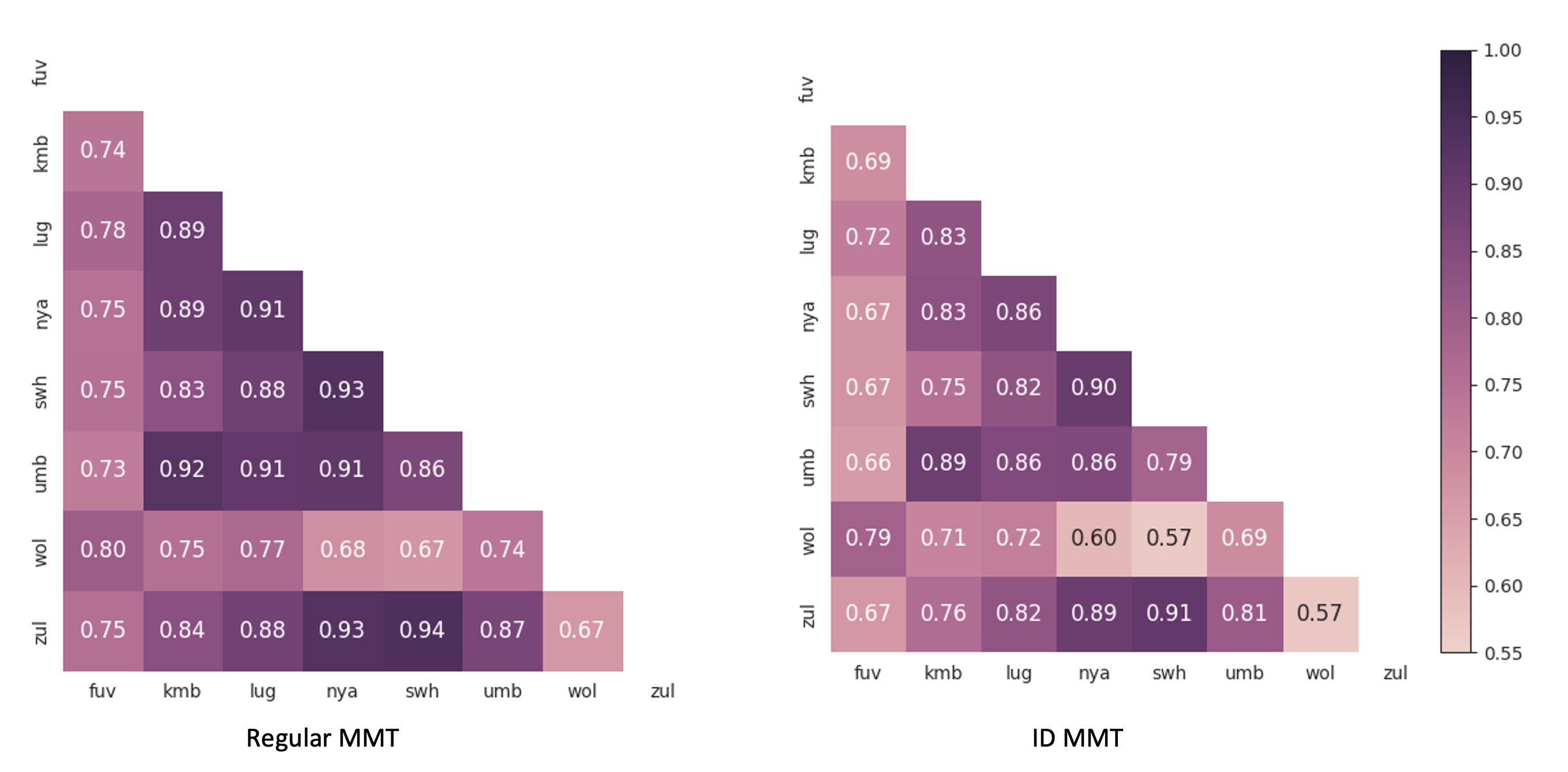}}
         \caption{90\% low-sensitive parameters}
         \label{fig:heatmap_low}
     \end{subfigure}
     \hfill
     \begin{subfigure}[b]{0.7\textwidth}
         \centering
         \resizebox{1\linewidth}{!}{
         \includegraphics[width=\textwidth]{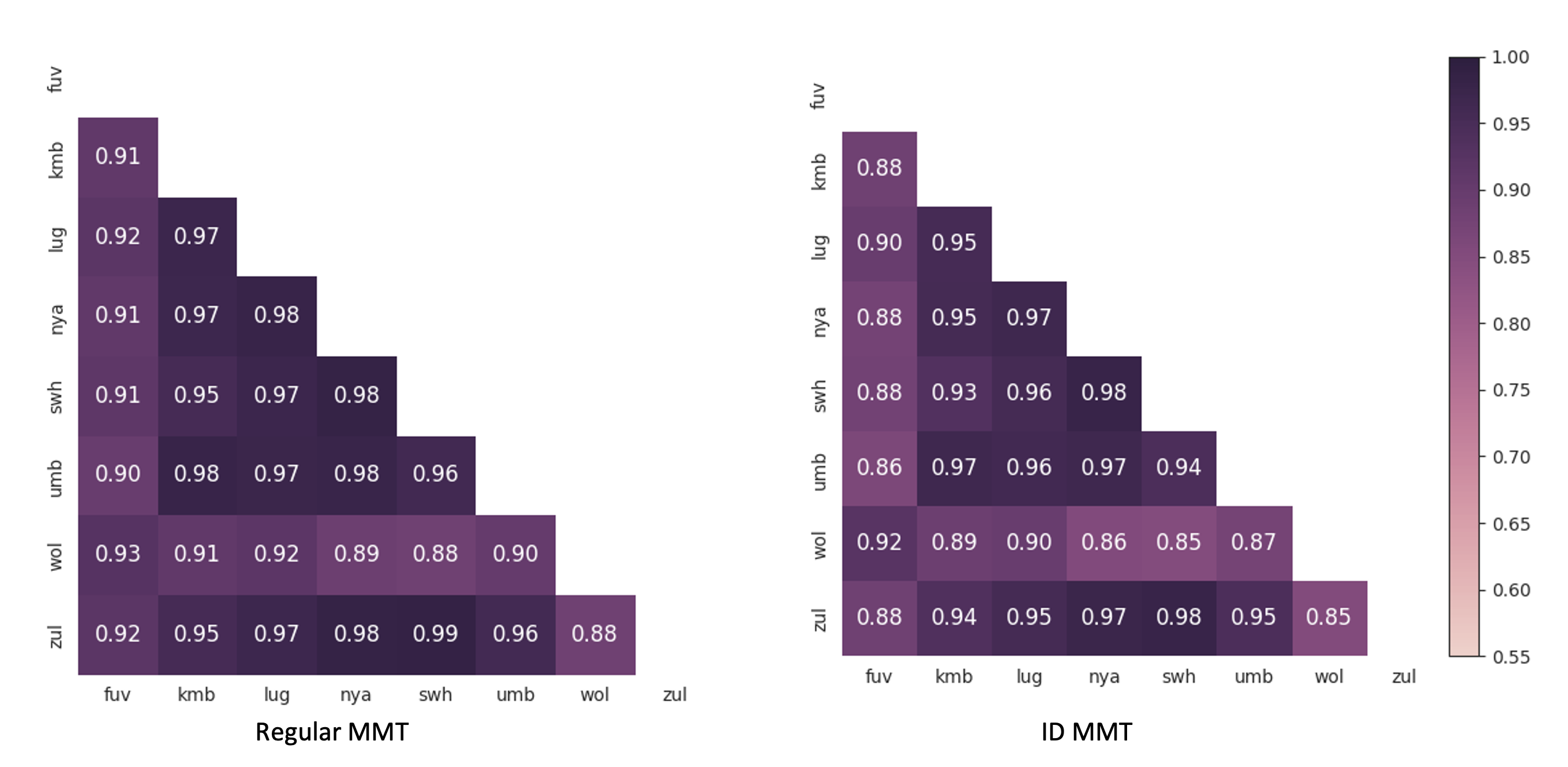}}
         \caption{10\% high-sensitive parameters}
         \label{fig:heatmap_high}
     \end{subfigure}
     \caption{PCC between the lists of parameter sensitivity of every language (left for regular MMT and right for MMT with ID). We show contribution similarity of two groups of parameters, i.e., top 10\% high-sensitive parameters and the remaining 90\% parameters. The lower score between two languages represents the less similarity of parameter contributions for these two languages, which means more contribution disagreements and parameters are more language-specific. } 
     \label{fig:heatmap}
\end{figure*}

Next, we study whether parameter contributions are more language-specific or just shareable across all languages after ID. Given the $i^\text{th}$ language $l_i$, we compute the sensitivities (Equation \ref{eq:score-define3}) of all parameters and flatten them into a list.  Then, we calculate the Pearson correlation coefficients (PCC) $p_{ij}$ between sensitivity lists of any arbitrary pair of languages $l_i$ and $l_j$. A lower $p_{ij}$ indicates that there are more contribution (sensitivity) disagreements between languages $l_i$ and $l_j$.  We plot a heat map to visualize $p_{ij}$ for every language pair. Taking into account that the top 10\% parameters usually dominate the contribution \citep{xiao2019autoprune,sanh2020movement}, we consider the performance of two groups of parameters, high-sensitive (top 10\% most sensitive) and low-sensitive (the remaining 90\%) parameters, respectively. Figure \ref{fig:heatmap} shows that all $p_{ij}$ in both groups become lower, indicating there is lower sensitivity similarity between different languages for the same parameters, which means the model becomes more language-specific after ID. For instance, sensitivity similarity between \texttt{zul} and \texttt{wol} drops from 0.67 to 0.57 in the low-sensitive group. However, the $p_{ij}$ of low-sensitive parameters drops much more than high-sensitive ones, and high-sensitive parameters still hold high similarity (over 0.9). Thus, low-sensitive parameters mostly have language-specific properties while high-sensitive parameters tend to play `language-agnostic' roles. Overall, parameters are more language-specific after ID\footnote{The overall parameter contribution is still more balanced as claimed in \citet{xu2022importance}. We leave further discussion on this to Appendix \ref{app:balanced}.}. In fact, learning more language-specific parameters through ID in MMT leads to better performance as seen in Section \ref{sec:why_id_exp}. These findings align with the results of recent studies which investigate language-specific parameters \citep{lin-etal-2021-learning,zhang2021share,nllb}, indicating the importance of language-specific parameters.

\begin{figure}[ht]
    \centering
    \resizebox{0.85\linewidth}{!}{
    \includegraphics[width=7.5cm]{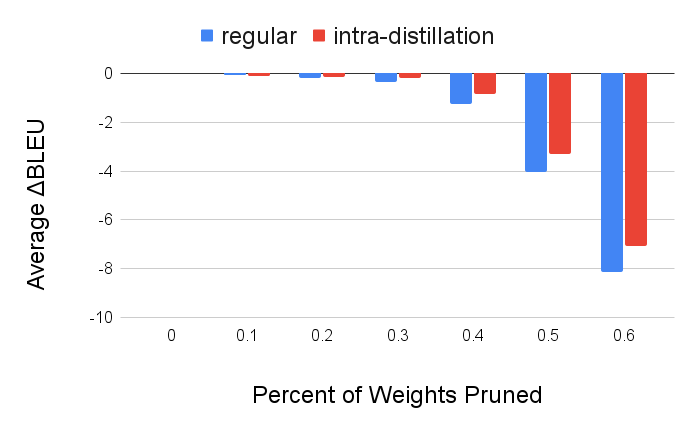}}
    \caption{Change in average \texttt{xxx}$\rightarrow$\texttt{eng} translation performance across 8 languages versus pruning ratio. Models are pruned starting with the least language-specific parameters. }
    \label{fig:prune}
\end{figure}

\subsection{The Importance of Language-Specific Parameters}
\label{sec:parameter_ls}
Here, we study the reason \textit{why language-specific parameters are important and how much they contribute}. To investigate this, we first measure the degree of language-specificity of all parameters based on Equation \ref{eq:ls}. We explore the contribution of language-specific parameters with respect to the BLEU scores. Then, we conduct one-shot unstructured pruning with respect to BLEU scores in order of the degree of language-specificity for both models with and without ID, starting with the least language-specific parameters\footnote{Note that, as shown in Figure \ref{fig:heatmap_high}, the 10\% most sensitive parameters are highly language-agnostic. They are easy to classify as less language-specific and can be pruned, but pruning them would lead to near-random performance (BLEU$\approx$0), making it hard to evaluate the importance of more language-specific parameters. Thus, we keep the top 10\% sensitive parameters and prune the rest of parameters that display a more language-specific behavior.}. As more parameters are pruned, a slower performance drop means that a higher contribution comes from the remaining more language-specific parameters. Figure \ref{fig:prune} shows the average BLEU drop across 8 languages versus the percentage of parameters pruned. After pruning the less language-specific parameters, the rest of the more language-specific parameters in the model with ID are able to preserve better performance, indicating the importance of more language-specific parameters.
%Pruning the least language-specific parameters is seen to have little effect on BLEU score, revealing the importance of language-specific parameters towards overall translation performance. Furthermore, we see how this effect is even more pronounced for the model with ID.
% In conclusion, we see that ID helps MMT models learn more language-specific parameters.

\section{Proposed Self-Supervision Method}
\label{sec:method}
We extend our study of language-awareness to MMT models co-trained with self-supervised objectives that have been shown to improve translation performance. We first propose a simple but effective self-supervised learning objective, \textbf{concurrent denoising} (CD), and then investigate the effectiveness of ID in helping improve multi-task optimization challenges of co-training CD and MMT tasks together by learning more language-specific parameters.

\subsection{Concurrent Denoising}
Self-supervised learning objectives usually involve sentence denoising either on the encoder side, such as MLM \citep{devlin2019bert}, or on the decoder side, such as DAE \citep{mbart}. Jointly denoising sentences on both the encoder and the decoder sometimes is better than a single denoising objective \citep{wang-etal-2020-multi,kim2021scalable} for MMT, but the training cost is doubled as we need to calculate the loss for the same monolingual sentence twice (masked in two different ways). We propose \textbf{concurrent denosing}, a self-supervised task that denoises a single masked sentence both on the encoder and decoder sides, which not only reduces the training time but also improves the language understanding of the model to result in better MMT performance.
\begin{figure}[ht]
    \centering
    \resizebox{1\linewidth}{!}{
    \includegraphics[width=7.5cm]{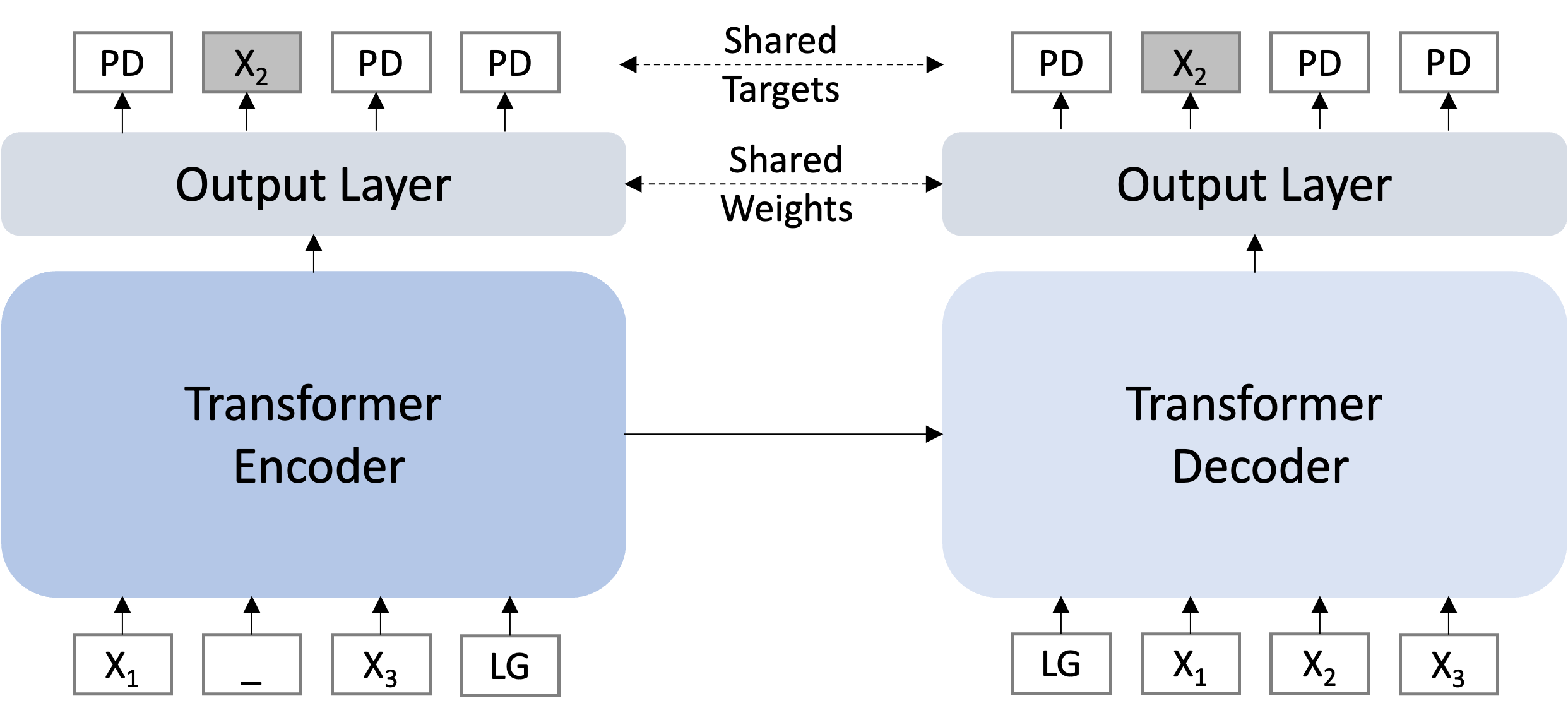}}
    \caption{Concurrent denoising. In the example input sentence `$\text{X}_1$ $\text{X}_2$ $\text{X}_3$', the token $\text{X}_2$ is masked. The encoder and decoder share the same output projection layer and target tokens to predict the masked token. We only calculate the loss for the masked token prediction. \texttt{PD} represents the target token loss padding and \texttt{LG} is a special language token.}
    \label{fig:cd}
\end{figure}

We add noise to the monolingual data by whole-word masking \citep{devlin2019bert}, where we randomly replace $r_m\%$ words with the special token \texttt{<mask>}. During the replacement process, each word has a 10\% chance not to be masked, and another 10\% chance to be replaced with other random tokens. The encoder and decoder use a shared output layer to reconstruct the original sentence. The loss for the encoder and decoder side are denoted as $\mathcal{L}_{e}$ and $\mathcal{L}_{d}$ respectively\footnote{Unlike DAE training on the decoder side, we zero out the losses which predict non-masked tokens.}. The total training loss combining translation loss $\mathcal{L}_{MMT}$ and two self-supervised losses is
\begin{equation}
    \mathcal{L} = \mathcal{L}_{MMT} + \mathcal{L}_{e} + \mathcal{L}_{d}.
    \label{eq:loss_simple}
\end{equation}
Concurrent denoising is illustrated in Figure \ref{fig:cd}. Two key differences between our concurrent denoising method and regular MLM or DAE methods are worth highlighting.

\paragraph{Shared Output Projection} 
Since the decoder has an output projection layer while the encoder does not, \citet{wang-etal-2020-multi} train the encoder with MLM by using an additional projection layer. However, we utilize the decoder projection layer as a shared layer for both encoder and decoder to reconstruct the sentence, which significantly reduces model parameters. This is because the projection layer is usually large when we have a large vocabulary size. We show the effect of using a shared projection layer in Appendix \ref{app:shared_layer}.

\paragraph{Shared Target Tokens} 
Since the output representations of the encoder and decoder are fed to the same projection layer, we want them to predict the same target token at the same position for the stability of the projection layer training. To achieve this, we carefully design our language token positions. Instead of only prepending a special language token at the beginning of the source sentence \citep{johnson-etal-2017-googles}, we append the special language token on the source side and also prepend it on the decoder side (As shown in Figure \ref{fig:cd}). This design also applies to MMT. In this way, we can avoid the encoder and decoder from predicting the same token at different positions.

\subsection{Concurrent Denoising with Intra-Distillation}
We investigate whether ID helps concurrent denoising to improve overall performance. We apply ID to the co-training of CD and MMT tasks. Following Equation  \ref{eq:id_final_loss} and \ref{eq:loss_simple}, our final loss is
\begin{multline}
     \mathcal{L} = \frac{1}{K}(\sum_{i=1}^K\mathcal{L}_{MMT_i} + \sum_{i=1}^K\mathcal{L}_{e_i} + \sum_{i=1}^K\mathcal{L}_{d_i}) + \\ \alpha (\mathcal{L}_{id\_MMT} + \mathcal{L}_{id\_e} + \mathcal{L}_{id\_d}),
    \label{eq:loss_final}
\end{multline}
where $\mathcal{L}_{id\_MMT}$, $\mathcal{L}_{id\_e}$ and $\mathcal{L}_{id\_d}$ respectively represent the ID loss for translation, encoder denoising and decoder denoising (i.e., $\mathcal{L}_{id\_e}$ minimizes the difference of the $K$ encoder outputs based on Equation \ref{eq:id_loss}, etc.). The $i$ index in $\mathcal{L}_{e_i}$, $\mathcal{L}_{MMT_i}$ and $\mathcal{L}_{d_i}$ indicates that these losses are for the $i^\text{th}$ forward pass.

\section{MMT+SSL Experiments}
\label{sec:exp}
\subsection{Baselines}
We consider three strong baselines. All baselines are our own implementation following the settings from the original papers.
\paragraph{DAE} \citet{nllb} learn the effects of the causal language modeling (LM) and DAE objectives \citep{mbart}. Since they find that DAE performs better than LM or LM+DAE, we only compare our methods with the DAE objective.
\paragraph{DAE+MLM} \citet{wang-etal-2020-multi} study a multi-task learning framework which jointly trains the MMT, MLM and DAE objectives, where MLM and DAE reconstruct sentences noised by different masking methods. \citet{kim2021scalable} also investigate the effectiveness of ELECTRA \citep{clark2019electra}. They conclude that DAE+MLM is better than DAE+ELECTRA.
\paragraph{MASS}  \citet{mass-mono} and \citet{siddhant2022towards} utilize MASS (masked sequence to sequence pre-training) \citep{mass} to improve the MMT performance. Similar to MLM which predicts masked tokens on the encoder side, MASS masks a fragment of a sentence and predicts the masked fragment but on the decoder side.

\begin{figure}[ht]
     \centering
     \begin{subfigure}[b]{0.4\textwidth}
         \centering
         \resizebox{1\linewidth}{!}{
         \includegraphics[width=\textwidth]{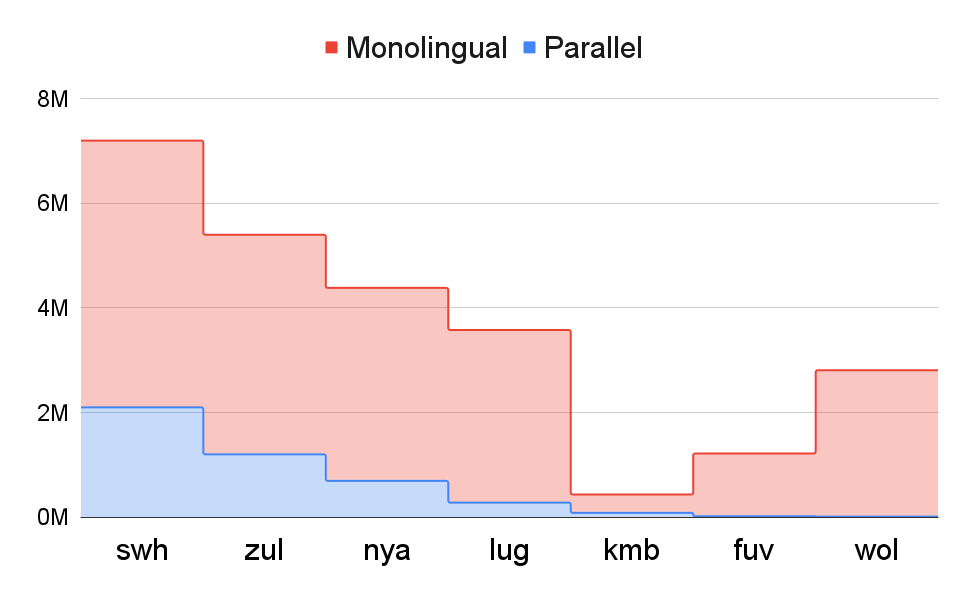}}
         \caption{M8 dataset}
         \label{fig:lg_stat_8}
     \end{subfigure}
     \hfill
     \begin{subfigure}[b]{0.4\textwidth}
         \centering
         \resizebox{1\linewidth}{!}{
         \includegraphics[width=\textwidth]{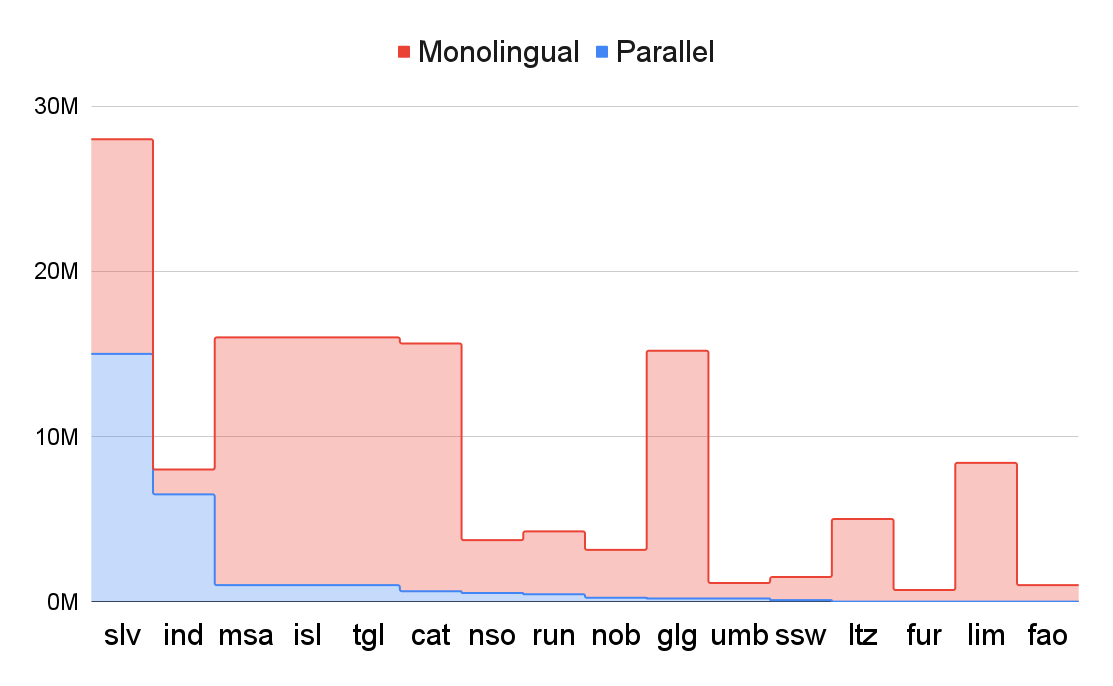}}
         \caption{M15 dataset}
         \label{fig:lg_stat_15}
     \end{subfigure}
     \caption{The statistics of monolingual and parallel data for M8 and M15 are presented. The languages are arranged in descending order of parallel data size. } 
     \label{fig:lg_stat}
\end{figure}
\subsection{Datasets}
In addition to the M8 dataset described in Section \ref{sec:why_id_exp}, we also build a larger dataset (M15), covering 15 languages. In composing this dataset, we take into account linguistic diversity and data size. The resulting dataset has languages from 6 linguistic families and a balanced number of high-resource, low-resource and very low-resource languages. Detailed information on this dataset is in Appendix \ref{app:m15_info}. We randomly sample at most 3M monolingual samples per language for M8, and 15M for M15. The distribution of monolingual data and parallel data for M8 and M15 is shown in Figure \ref{fig:lg_stat}. Note that we also use parallel data for self-supervised learning, so the true monolingual data size includes bitext data. We use the \textsc{Flores-200} dataset for evaluation. All datasets come from the primary bitext and monolingual data used for the NLLB-200 model \citep{nllb}.

% \begin{figure}[ht]
%     \centering
%     \resizebox{1\linewidth}{!}{
%     \includegraphics[width=7.5cm]{figures/lg_stat.png}}
%     \caption{The statistic information of monolingual data and parallel data for M8 and M15. The total 23 languages are sorted decreasingly by their parallel data size.}
%     \label{fig:lg_stat}
% \end{figure}

\subsection{Data Sampling}
We use a data sampling temperature of $T=1$ suggested by \citet{nllb} to train on the MMT objective. For monolingual data, we use a temperature of $\frac{10}{7}$ to balance the SSL training, as suggested by \citet{mbart}. During co-training, we mix the two sources in an equal ratio (50\% monolingual data (including bitext used for SSL training) with self-supervision and 50\% parallel data).

% .\footnote{We use a lower value compared to \citet{aharoni2019massively}'s $T=5$, which we noticed led to low-resource languages over-fitting and high-resource languages under-fitting. We hypothesize that the domain of \textsc{Flores-200} differs from that of the training data,  so overfitting is more easily detected compared to other datasets.}
\subsection{Training and Evaluation Details}
All experiments consider both the \texttt{eng}$\rightarrow$\texttt{xxx} and \texttt{xxx}$\rightarrow$\texttt{eng} directions and use the Transformer architecture \citep{transformer}. We use Transformer$_\text{big}$ (242M parameters, 6 layers, 16 heads, 1,024 hidden dimension, 4,096 FFN dimension) for M8 experiments. For M15 experiments, we double the layers of Transformer$_\text{big}$ (418M parameters). We use a vocabulary of size 32k for both M8 and M15 with SentencePiece \citep{sentencepiece}. The batch size is 30K tokens. We warm-up for the first 8K steps. We set the total training steps to 100K and 300k for M8 and M15 respectively, with patience set to 10 for early stopping. We forward pass the model twice ($K$=2) to conduct ID. We set the ID weight $\alpha=5$. During concurrent denoising, the masking ratio is set to $r_m=30\%$. We also show the effect of masking ratio in Appendix \ref{app:mask_ratio}. During generation, we use beam search with a beam size of 5 and a length penalty of 1.0. All models are evaluated with sacreBLEU (\texttt{spm} tokenizer).

\begin{table}[t]
\centering
\resizebox{1\linewidth}{!}{
\begin{tabular}{l|cccc}
\hline
Method        & High           & Low            & Very Low      & All            \\ \hline
\textit{M8 results} &&&&                                         \\
Regular MMT   & 31.70          & 12.57          & 6.92          & 15.94          \\
+DAE \citep{nllb}        & 32.69          & 13.38          & 7.57          & 16.75          \\
+DAE+MLM \citep{wang-etal-2020-multi}     & 33.05          & 13.93          & 8.20          & 17.27          \\
+MASS  \citep{mass-mono}     & 32.64          & 13.03          & 6.79          & 16.37          \\
+CD (ours)    & 32.92          & 13.94          & 8.38          & 17.29          \\
+CD+ID (ours) & \textbf{35.16} & \textbf{15.18} & \textbf{9.18} & \textbf{18.69} \\ \hline
\textit{M15 results} &&&&                                         \\
Regular MMT   & \textbf{39.87} & 35.20          & 24.45         & 33.17          \\
+DAE \citep{nllb}         & 38.46          & 34.05          & 26.23         & 32.91          \\
+DAE+MLM \citep{wang-etal-2020-multi}     &   38.60     & 34.00   &    25.49    & 32.70            \\
+MASS  \citep{mass-mono}     & 38.53          & 33.93         & 22.79        & 31.75         \\
+CD (ours)    & 39.23          & 34.88          & 28.21         & 34.11          \\
+CD+ID (ours) & 39.58          & \textbf{35.53} & \textbf{29.43} & \textbf{34.85}          \\ \hline
\end{tabular}
}
\caption{Overall \texttt{xxx}$\rightarrow$\texttt{eng} BLEU for M8 and M15.}
\label{tab:xx-eng-main}
\end{table}

\begin{table}[ht]
\centering
\resizebox{1\linewidth}{!}{
\begin{tabular}{l|cccc}
\hline
Method        & High           & Low            & Very Low & All            \\ \hline
\textit{M8 results} &&&&                                   \\
Regular MMT   & 34.14          & 11.47          & 5.75     & 15.71          \\
+DAE  \citep{nllb}        &    34.35       &    11.41       & 5.79     &    15.74       \\
+DAE+MLM \citep{wang-etal-2020-multi}      &    34.48   &  11.45   &5.20      &15.64                \\
+MASS  \citep{mass-mono} & 34.02 & 11.53 & 4.75 & 15.46 \\
+CD (ours)    & 34.87          & 11.50          & \textbf{5.90}     & 15.94          \\
+CD+ID (ours) & \textbf{35.83} & \textbf{11.90}  & 5.81   & \textbf{16.37} \\ \hline
\textit{M15 results} &&&&                               \\
Regular MMT   &  \textbf{38.44}       & 31.62 & 16.46    & 28.84         \\
+DAE  \citep{nllb} &  37.46      & 30.86   & 18.39           & 28.90             \\
+DAE+MLM \citep{wang-etal-2020-multi}     & 37.99   & 30.98        &18.05    & 29.01                \\
+MASS  \citep{mass-mono} & 38.19& 31.20 & 17.88 & 29.09 \\
+CD (ours)    & 37.74   & 30.94    & 19.04  & 29.24                \\
+CD+ID (ours) &  38.29 & \textbf{31.71}    & \textbf{19.43}    & \textbf{29.81}                \\ \hline
\end{tabular}
}
\caption{Overall \texttt{eng}$\rightarrow$\texttt{xxx} BLEU for M8 and M15.}
\label{tab:eng-xx-main}
\end{table}

\subsection{Results}
\label{sec:results}
The overall results for the \texttt{xxx}$\rightarrow$\texttt{eng} and \texttt{eng}$\rightarrow$\texttt{xxx} directions are shown in Tables \ref{tab:xx-eng-main} and \ref{tab:eng-xx-main}. For both M8 and M15, and both translation directions, concurrent denosing is better than all aforementioned baselines, and combining it with ID further improves upon the baselines by an even larger margin. For instance, our method outperforms MASS by 11.3\% and 3.7\% on M8 and M15 respectively, averaged across all languages and directions. We also show the effectiveness of ID on other objectives like DAE in Section \ref{sec:ablation}, but the results are subpar compared to CD+ID. 

Aligned with the findings of \citet{wang-etal-2020-multi,kim2021scalable}, we observe that DAE+MLM is better than DAE alone in M8 \texttt{xxx}$\rightarrow$\texttt{eng}, but the improvements become very minor when it comes to M8 \texttt{eng}$\rightarrow$\texttt{xxx} or when scaling to 15 languages. MASS performs similarly or better than DAE in the \texttt{eng}$\rightarrow$\texttt{xxx} but worse in the \texttt{xxx}$\rightarrow$\texttt{eng}.

% In the M15 experiments, the performance of high-resource languages is slightly worse with any SSL method when compared to the MMT only baseline. With our approach, however, the overall performance of other language categories shows significant improvement, which is similar to the observations of \citet{nllb}. However, this phenomenon with high-resource languages does not occur on M8. We hypothesize that, since M8 is a smaller dataset compared to M15, the model capacity is sufficient to learn from additional monolingual data to improve even high-resource languages.

% In the M15 experiments, high-resource languages perform slightly worse with SSL methods compared to the MMT only baseline. However, our approach improves the overall performance of other language categories, similar to the observations of \citet{nllb}. The phenomenon with high-resource languages does not occur on M8, possibly due to the smaller dataset size allowing for sufficient model capacity to learn from additional monolingual data.

In M15, high-resource languages perform slightly worse with SSL methods compared to the MMT only baseline, but improves other categories, similar to the observations of \citet{nllb}. It does not occur on M8, possibly due to the smaller dataset size allowing for sufficient model capacity to learn from additional monolingual data.

Note that the effectiveness of SSL such as DAE and MASS is not as pronounced as reported by \citet{wang-etal-2020-multi} and \citet{siddhant2022towards}. However, it is necessary to consider the for domain mismatch between the training and evaluation data. As demonstrated by \citet{siddhant2022towards}, a significant decline in performance can occur when either monolingual or bitexts diverge from the evaluation domain. In our study, the training data is sourced from NLLB-200 and FLORES-200, which encompasses a wide range of domains. we hypothesize that this contributes to the observed lessened effectiveness of SSL techniques in our experiments.

\section{Analysis}

\subsection{Ablation Study}
\label{sec:ablation}
The final loss, described in Equation \ref{eq:loss_final}, has 6 loss terms.  Except for the translation loss, we ablate the relative contribution of all the other 5 loss terms to the translation task performance. In Table \ref{tab:ablation_train}, we show the results of this ablation study on M8 \texttt{xxx}$\rightarrow$\texttt{eng} directions. Method \circled{1} is the regular MMT model and method \circled{2} is ID training only for MMT (the same result as in Section \ref{sec:why_id_exp}). Method \circled{3} is the same as the MMT+DAE method. With the help of ID for the decoder denoising (method \circled{4}) and an additional ID for translation (method \circled{5}), translation performance can respectively obtain +0.41 and +0.98 BLEU on average compared to \circled{3}. Note that method \circled{5} is the MMT+DAE+ID method. Compared to our MMT+CD+ID method, it substantially underperforms our method (17.73 vs. 18.69), which shows that our method could better stimulate the potential of ID.  The results for methods \circled{6}, \circled{7} and \circled{8} indicate the effectiveness of encoder denoising with CD and applying ID. Overall, the translation performance improves by including all the loss terms.

\begin{table}[ht]
\centering
\resizebox{1\linewidth}{!}{
\begin{tabular}{llc}
\hline
 & Method                          & Avg. BLEU \\ \hline
\circled{1}&  $\mathcal{L}_{MMT}$                         &    15.94     \\
\circled{2}&$\mathcal{L}'_{MMT} + \mathcal{L}_{id\_MMT}$                    &  17.15       \\ 
\circled{3} &$\mathcal{L}_{MMT} + \mathcal{L}_{d}$                           & 16.75        \\
\circled{4}&$\mathcal{L}'_{MMT}+\mathcal{L}'_{d}  +\alpha \mathcal{L}_{id\_d}$                    &  17.16       \\
\circled{5}&$\mathcal{L}'_{MMT}+\mathcal{L}'_{d}  +\alpha (\mathcal{L}_{id\_d}+ \mathcal{L}_{id\_MMT})$                    &  17.73       \\
\circled{6}&$\mathcal{L}_{MMT} + \mathcal{L}_e + \mathcal{L}_d$                          &   17.29      \\
\circled{7}&$\mathcal{L}'_{MMT} + \mathcal{L}'_e + \mathcal{L}'_d + \alpha (\mathcal{L}_{id\_d} + \mathcal{L}_{id\_e})$          &    17.59      \\
\circled{8}&$\mathcal{L}'_{MMT} + \mathcal{L}'_e + \mathcal{L}'_d + \alpha (\mathcal{L}_{id\_d} + \mathcal{L}_{id\_e} + \mathcal{L}_{id\_MMT})$ &  \textbf{18.69}      \\ \hline
\end{tabular}
}
\caption{Ablation study on loss terms. For simplicity, we use $\mathcal{L}'$ to represent the mean loss of $K$ forward pass, e.g., $\mathcal{L}'_{e} = \frac{1}{K}\sum_{i=1}^K\mathcal{L}_{e_i}$.}
\label{tab:ablation_train}
\end{table}

\begin{figure*}[ht]
    \centering
    \resizebox{0.7\linewidth}{!}{
    \includegraphics[width=7.5cm]{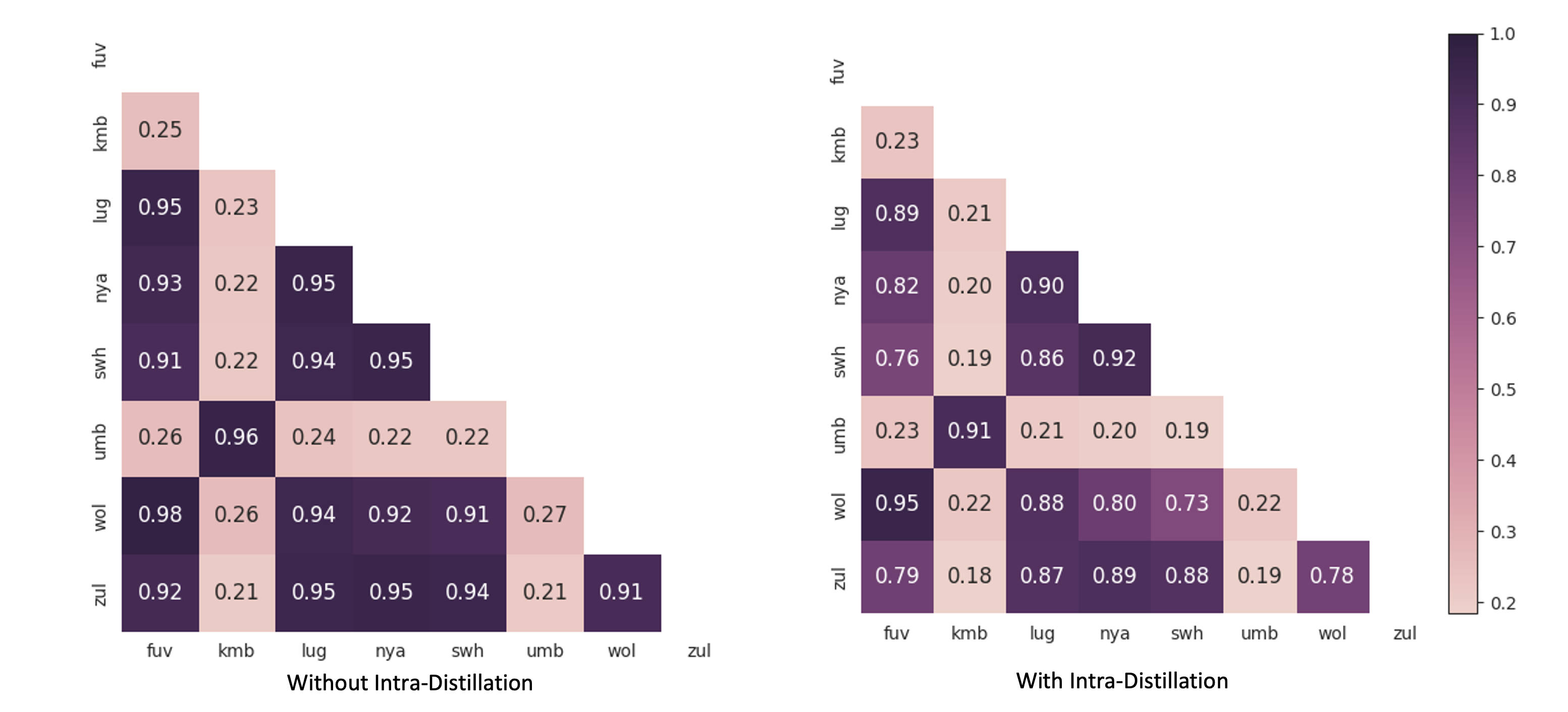}}
    \caption{Parameter contribution similarity among all language pairs, evaluated by PCC for the CD task before (left) and after (right) ID.}
    \label{fig:heat_ssl}
\end{figure*}

\begin{figure}[ht]
    \centering
    \resizebox{1\linewidth}{!}{
    \includegraphics[width=7.5cm]{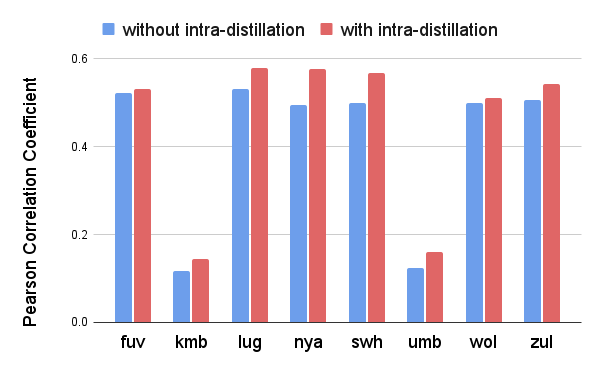}}
    \caption{Parameter contribution similarity between MMT and CD for each language with and without ID. }
    \label{fig:para_sim}
\end{figure}
\subsection{Language-Specific Parameters for SSL}
% In Section \ref{sec:why_id}, we observed that intra-distillation helps MMT learn more language-specific parameters and improve the model generalization. We are also interested whether the model also learns more language-specific parameters for the SSL task (here we investigate concurrent denoising). We still take M8 \texttt{xxx}$\rightarrow$\texttt{eng} as an example to study the two question. We draw a heatmap to illustrate the Pearson correlation coefficient between every language pairs in Figure \ref{fig:heat_ssl}. As expected, parameter sensitivity similarity becomes lower for all languages, which means there are more language-specific parameters when we train SSL methods with ID.

In Section \ref{sec:why_id}, we observed that ID helps MMT learn more language-specific parameters and improve model generalization. We are also interested in understanding 1) whether the model also learns more language-specific parameters for the SSL task (here we investigate CD), and 2) what is the relationship of parameter contribution between MMT and SSL tasks for the same language. We use the \texttt{xxx}$\rightarrow$\texttt{eng} direction of the M8 dataset as an example to study these questions.

In Figure \ref{fig:heat_ssl}, we plot a heat map to illustrate the PCC of all parameter sensitivities between every language pair. As expected, parameter sensitivity similarity becomes lower for all languages, which means there are more language-specific parameters when we train SSL methods with ID. For the second question, in Figure \ref{fig:para_sim} we show the parameter sensitivity similarity between the MMT and CD tasks for each language. The contribution similarity becomes higher between the two tasks for every language with ID. This is expected, since the losses of MMT and CD have the same objective on the decoder side, i.e., text generation conditioned on another text. This is also another reason why SSL tasks can help multilingual translation.
%\footnote{This may be also the reason why we do not notice parameters are more MMT-specific or CD-specific, though we claim that intra-distillation could help model learn task- (language-) specific parameters. We hypothesize that the main difference that model recognizes different tasks is different language inputs (tokens, embeddings and scripts), so it mainly learns langugae-specifc parameters.}.

\section{Conclusions}
% In this work, we first propose a measure method to quantify the language specificity degree for all parameters, and thorough analysis showing that intra-distillation helps multilingual translation models by learning more language-specific parameters, and demonstrate the importance of language-specific parameters. Next, We propose concurrent denoising, a complementary task that co-trains with the MMT task, which beats mutiple state-of-the-art SSL methods in improving MMT performance. Moreover, we apply intra-distillation to the co-training scheme and show that combining these two methods offers a further substantial improvement.

We show extensive analysis that intra-distillation training helps multilingual translation by learning more language-specific parameters. We propose concurrent denoising, improving upon multiple state-of-the-art self-supervised learning methods. Moreover, we demonstrate that applying intra-distillation to the above co-training scheme offers further improvements to translation performance.

\section*{Limitations}
Although we show improvements using our methods on multiple languages from diverse language families on multilingual machine translation, it should be noted that the generalizability of our findings to other multi-task learning settings, such as those involving the combination of tasks such as named entity recognition, part-of-speech tagging, and question answering, remains uncertain. This is due to the fact that our study primarily focused on the utilization of intra-distillation to learn task-specific parameters on multilingual machine translation and did not investigate the aforementioned tasks. Furthermore, with intra-distillation we need to perform more than one forward pass, leading to a trade-off between higher performance and increased training time -- which, for many use-cases, could be arguably acceptable.

\section*{Acknowledgements}
We would like to thank anonymous reviewers for their valuable comments. We also thank Alex Guo, Simeng Sun, and Weiting Tan for their helpful suggestions.

\bibliography{anthology,custom}
\bibliographystyle{acl_natbib}

\clearpage
\appendix

\section{Analysis of Intra-Distillation for \texttt{eng}$\rightarrow$\texttt{xxx}}
\label{app:eng-xx}

\begin{table}[ht]
\centering
\resizebox{1\linewidth}{!}{
\begin{tabular}{l|cccc}
\hline
Method             & High  & Low   & Very Low & All   \\ \hline
Regular            & 34.14 & 11.47 & \textbf{5.75}     & 15.71 \\
Intra-Distillation & \textbf{35.05} & \textbf{13.79} & 5.69     & \textbf{16.07} \\ \hline
\end{tabular}
}
\caption{M8 \texttt{eng}$\rightarrow$\texttt{xx} results of regular MMT and MMT with intra-distillation.}
\label{tab:id_vs_regular_eng_xx}
\end{table}

\begin{figure}[ht]
    \centering
    \resizebox{1\linewidth}{!}{
    \includegraphics[width=7.5cm]{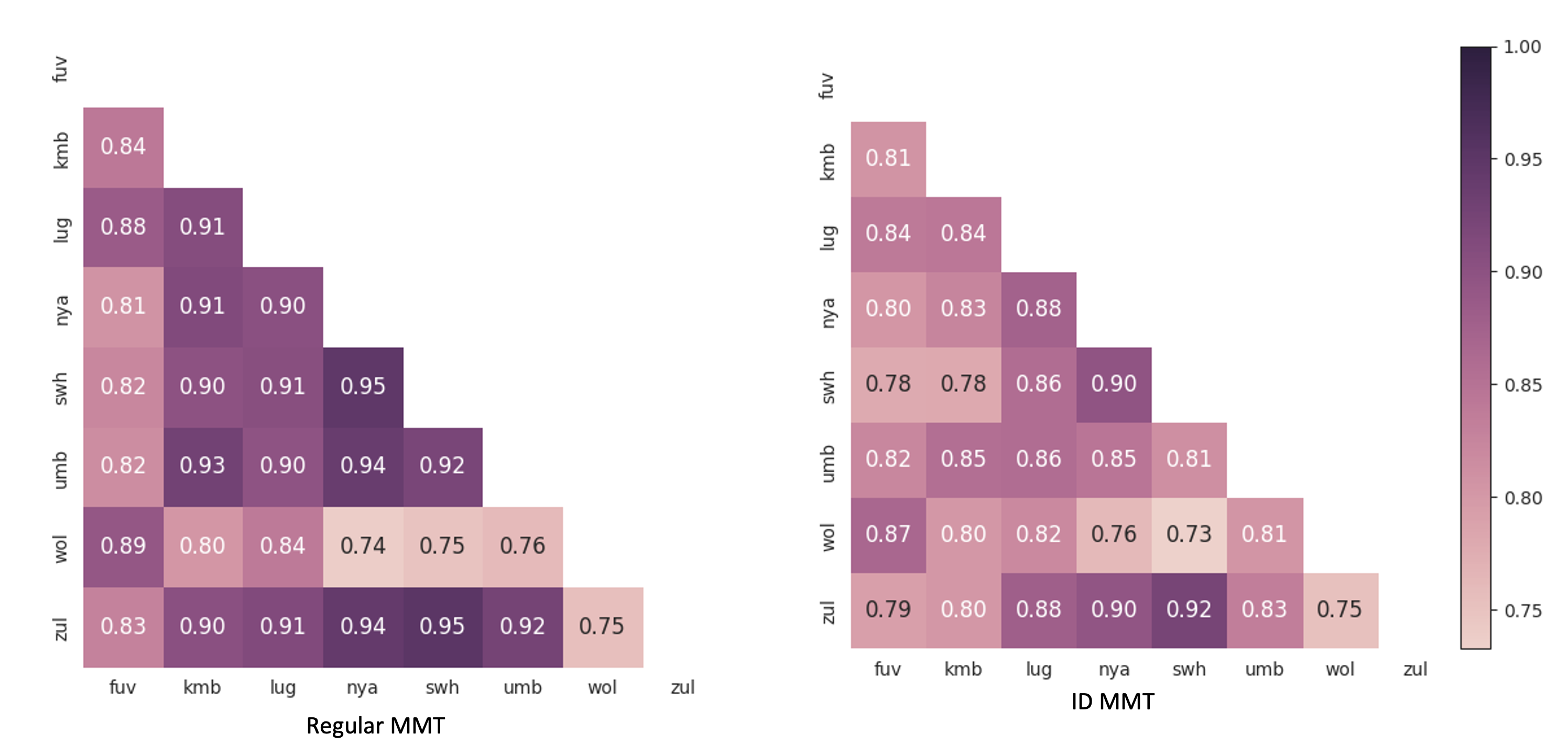}}
    \caption{PCC between the list of all parameter sensitivities across every language in the M8 \texttt{eng}$\rightarrow$\texttt{xxx} experiments. We compare the similarity between MMT with and without intra-distillation. }
    \label{fig:heat_eng_xx}
\end{figure}

Similar to Section \ref{sec:why_id}, the model with intra-distillation outperforms the regular MMT model by a large margin in the \texttt{eng}$\rightarrow$\texttt{xxx} direction, as shown in Table \ref{tab:id_vs_regular_eng_xx}. We still use a heat map to visualize the PCC of parameter sensitivity lists among every language pair in the \texttt{eng}$\rightarrow$\texttt{xxx} direction. In Figure \ref{fig:heat_eng_xx}, we show that contribution similarity becomes lower as well, which means that the model also learns more language-specific parameters. 

We also evaluate the importance of these language-specific parameters by following the same settings in Section \ref{sec:parameter_ls}. We conduct one-shot unstructured pruning, starting with the least language-specific parameters. We again see that the average BLEU scores of 8 languages from the model trained with intra-distillation drop slower after more parameters are pruned, indicating that these language-specific parameters learned by intra-distillation are able to preserve more performance.

\begin{figure}[ht]
    \centering
    \resizebox{0.85\linewidth}{!}{
    \includegraphics[width=7.5cm]{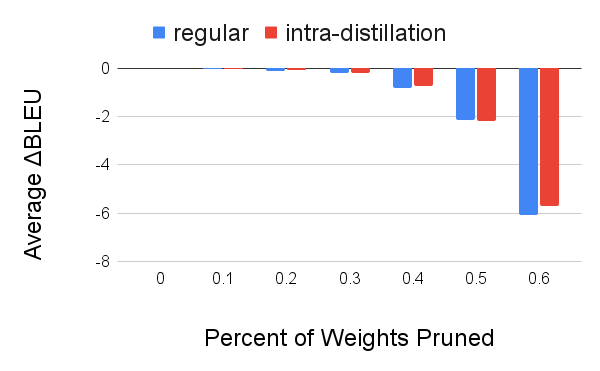}}
    \caption{Change of model performance averaged across 8 languages against increasing pruning ratio for the \texttt{eng}$\rightarrow$\texttt{xxx} translation task. Models are pruned starting with the least language-specific parameters. }
    \label{fig:prune_eng_xx}
\end{figure}

\section{More Balanced Parameter Contribution}
\label{app:balanced}
\begin{figure}[ht]
    \centering
    \resizebox{0.85\linewidth}{!}{
    \includegraphics[width=7.5cm]{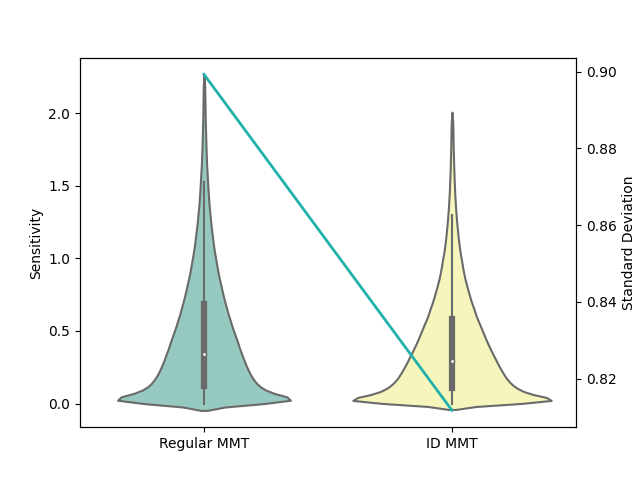}}
    \caption{Sensitivity distribution (violin plots aligned with left y-axis) along with their standard deviation (green curve aligned with right y-axis, lower means more balanced parameter contribution). Note that we also remove the top 1\% highest-sensitive parameters to ease the illustration. }
    \label{fig:dis_compare_all}
\end{figure}

We compute the sensitivity of all parameters by feeding a set of batches $\mathcal{B}$ that contains all language data in the M8 \texttt{xxx}$\rightarrow$\texttt{eng} experiment. We illustrate parameter sensitivity distribution in Figure \ref{fig:dis_compare_all}. Aligned with the findings in \citet{xu2022importance}, the distribution of parameter sensitivity becomes more balanced after using ID.

\section{Ablation Study on Shared Projection Layer}
\label{app:shared_layer}
Since we use a shared projection layer for both encoder and decoder denoising as well as for translation to reduce the model size and save memory, we investigate whether this sharing leads to  a performance drop. We conduct experiments on M8 \texttt{xxx}$\rightarrow$\texttt{eng} dataset. Table \ref{tab:ablation_shared_layer} shows that our method with shared layer slightly outperforms the one with separate output projection layers on average.

\begin{table}[ht]
\centering
\resizebox{1\linewidth}{!}{
\begin{tabular}{l|cccc}
\hline
Method             & High  & Low   & Very Low & All   \\ \hline
CD+ID (shared layer)            & \textbf{35.16} & \textbf{15.18} & 9.23     & \textbf{18.69} \\
CD+ID (NOT shared layer)  & 35.09 & 15.04 & \textbf{9.28}    & 18.61 \\ \hline
\end{tabular}
}
\caption{Comparison of concurrent denoising + intra-distillation with and without using a shared projection layer.}
\label{tab:ablation_shared_layer}
\end{table}

\section{M15 Language Information}
\label{app:m15_info}
We give a full account of the 15 languages in the M15 dataset in Table \ref{tab:m15_info}.
\begin{table*}[ht]
\centering
\resizebox{1\linewidth}{!}{
\begin{tabular}{lccccc}
\hline
Language           & \multicolumn{1}{c}{Language id} & \multicolumn{1}{c}{Parallel Data Size} & Resource Level & \multicolumn{1}{c}{Language family} & \multicolumn{1}{c}{Monolingual Data Size} \\ \hline
Northern Sotho     & nso                             & 526K                                   & Low          & Central Narrow Bantu                & 3.2M                                      \\
Rundi              & run                             & 454K                                   & Low          & Central Narrow Bantu                & 3.8M                                      \\
Swati              & ssw                             & 94K                                    & Very Low     & Central Narrow Bantu                & 1.4M                                      \\
Indonesian         & ind                             & 6.5M                                   & High         & Malayio-Polynesian                             & 1.5M                                      \\
Malay              & msa                             & 1M                                     & High         & Malayio-Polynesian                             & 15M                                       \\
Tagalog            & tgl                             & 1M                                     & High         & Malayo-Polinesian                 & 15M                                       \\
Bokmål (Norwegian) & nob                             & 238K                                   & Low          & North Germanic                      & 2.9M                                      \\
Icelandic          & isl                             & 1M                                     & High         & North Germanic                      & 15M                                       \\
Faroese            & fao                             & 4K                                     & Very Low     & North Germanic                      & 1.2M                                      \\
Slovene            & slv                             & 15M                                    & High         & Southwestern Slavic                 & 13M                                       \\
Luxembourgish      & ltz                             & 8K                                     & Very Low     & Western Germanic                    & 5M                                        \\
Limburgish         & lim                             & 5K                                     & Very Low     & Western Germanic                    & 8.4M                                      \\
Catalan            & cat                             & 634K                                   & Low          & Western Romance                     & 15M                                       \\
Galician           & glg                             & 195K                                   & Low          & Western Romance                     & 15M                                       \\
Friulian           & fur                             & 6K                                     & Very Low     & Western Romance                     & 730K                                     \\ \hline
\end{tabular}
}
\caption{The information of 15 languages in M15 dataset.}
\label{tab:m15_info}
\end{table*}

\section{Effect of Masking Ratio}
\label{app:mask_ratio}
\begin{figure}[ht]
    \centering
    \resizebox{1\linewidth}{!}{
    \includegraphics[width=7.5cm]{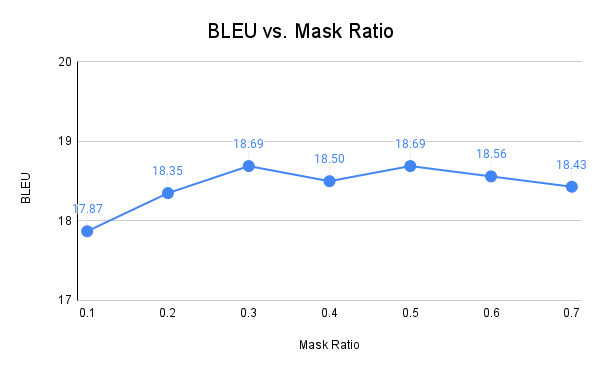}}
    \caption{MMT performance change along with masking ratio on the MMT+CD+ID method.}
    \label{fig:mask_ratio}
\end{figure}

We take MMT+CD+ID as our study case to investigate the effect of masking ratio $r_m\%$ on the MMT performance. We conduct experiments on M8 \texttt{xxx}$\rightarrow$\texttt{eng}. Figure \ref{fig:mask_ratio} shows that there is no big performance change when we set mask ratio between 0.3 and 0.6.

\end{document}